\documentclass{article}
\usepackage{spconf,amsmath,graphicx}

\usepackage{amsfonts}
\usepackage{caption}
\usepackage[labelformat=simple]{subcaption}
\usepackage{float}
\usepackage{amssymb}
\usepackage{enumitem}
\usepackage{algorithm} 
\usepackage{algorithmic}
\usepackage[algo2e]{algorithm2e}
\usepackage{amsthm,mathtools}
\usepackage{url}
\usepackage{stmaryrd}
\usepackage{cite}
\usepackage{multirow}
\usepackage[table,xcdraw]{xcolor}
\usepackage{url}
\usepackage{ragged2e}
\usepackage{epstopdf}

\newcommand{\etal}{\textit{et al}. }
\newcommand{\ie}{\textit{i}.\textit{e}., }

\DeclareMathOperator{\diag}{\operatorname{diag}}

\title{Hypergraph Convolutional Networks for Weakly-Supervised Semantic Segmentation}
\name{Jhony H. Giraldo$^{\S}$ \ Vincenzo Scarrica$^{\dagger}$ \ Antonino Staiano$^{\dagger}$ \ Francesco Camastra$^{\dagger}$ \ Thierry Bouwmans$^{\star}$}
\address{$^{\S}$ LTCI, Télécom Paris - Institut Polytechnique de Paris, France \\
         $^{\dagger}$ Dipartimento di Scienze e Tecnologie, Università di Napoli Parthenope, Italy\\
         $^{\star}$ Laboratoire Mathématiques, Image et Applications (MIA), La Rochelle Université, France}

\begin{document}
%
\maketitle

\begin{abstract}
Semantic segmentation is a fundamental topic in computer vision.
Several deep learning methods have been proposed for semantic segmentation with outstanding results.
However, these models require a lot of densely annotated images.
To address this problem, we propose a new algorithm that uses HyperGraph Convolutional Networks for Weakly-supervised Semantic Segmentation (HyperGCN-WSS).
Our algorithm constructs spatial and k-Nearest Neighbor (k-NN) graphs from the images in the dataset to generate the hypergraphs.
Then, we train a specialized HyperGraph Convolutional Network (HyperGCN) architecture using some weak signals.
The outputs of the HyperGCN are denominated pseudo-labels, which are later used to train a DeepLab model for semantic segmentation.
HyperGCN-WSS is evaluated on the PASCAL VOC 2012 dataset for semantic segmentation, using scribbles or clicks as weak signals.
Our algorithm shows competitive performance against previous methods.
\end{abstract}
\section{Introduction}
\label{sec:introduction}

Semantic segmentation is an important task in computer vision with multiple applications in image, 3D, and video processing \cite{ronneberger2015u,li2019deepgcns,giraldo2020graph}.
The main objective of semantic segmentation is to classify all the pixels in the images into some predefined classes.
Deep learning models have dominated the study of semantic segmentation in recent years \cite{long2015fully,badrinarayanan2017segnet,chen2017deeplab}.
However, these deep learning methods are usually very complex models containing millions of learnable parameters, and thus they require a lot of densely annotated images to perform well \cite{giraldo2020graph,giraldo2020semi}.

Currently, there is an increasing interest in weakly supervised learning \cite{zhou2018brief}, where the predictions are obtained with a limited amount of labels.
As a result, several studies have proposed Weakly-supervised Semantic Segmentation (WSS) methods \cite{xing2016weakly,lin2016scribblesup,pu2018graphnet,zhang2021affinity}, where graphical models have played a central role.
Particularly, Graph Convolutional Networks (GCNs) have been widely explored in WSS, reaching state-of-the-art performances \cite{pu2018graphnet,zhang2021affinity}.
However, these methods have focused on constructing graphs from individual images using spatial information.
Thus, these models waste crucial information that can be obtained from other images in the dataset.

\begin{figure}
    \centering
    \includegraphics[width=0.4\textwidth]{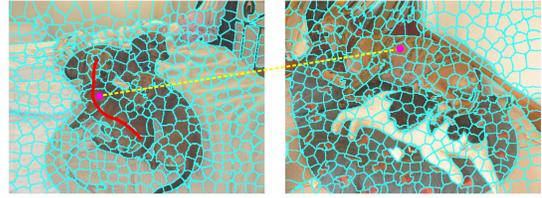}
    \caption{The idea of HyperGCN-WSS is to rely both on the spatial and structural information in the datasets.}
    \vspace{-0.32cm}
    \label{fig:idea_hypergcn}
\end{figure}

In this work, we propose a new algorithm named HyperGraph Convolutional Networks for Weakly-supervised Semantic Segmentation (HyperGCN-WSS), using scribbles and clicks as weak signals.
The key idea of our algorithm is to rely on spatial information as in \cite{pu2018graphnet,zhang2021affinity}, as well as on structural information that can be captured from other instances in the dataset.
Our algorithm uses HyperGCNs \cite{bai2021hypergraph} to capture such information.
HyperGCN-WSS is composed of 1) superpixel segmentation \cite{achanta2012slic} for node representation, 2) VGG-16 for feature extraction \cite{simonyan2015very}, 3) spatial and k-NN graph construction, 4) HyperGCN \cite{bai2021hypergraph} to generate pseudo-labels, and 5) DeepLabV3+ \cite{chen2018encoder} for semantic segmentation using the pseudo-labels as the ground-truth.
Fig. \ref{fig:idea_hypergcn} shows the motivation of HyperGCN-WSS, where one labeled superpixel (with a scribble) is connected to another non-labeled superpixel in the dataset, allowing the propagation of information between instances in the dataset.
HyperGCN-WSS is evaluated in the PASCAL VOC 2012 dataset \cite{everingham2015pascal} for semantic segmentation using scribbles and clicks as weak signals.
Our algorithm shows competitive performance against previous methods.

\begin{figure*}
    \centering
    \includegraphics[width=\textwidth]{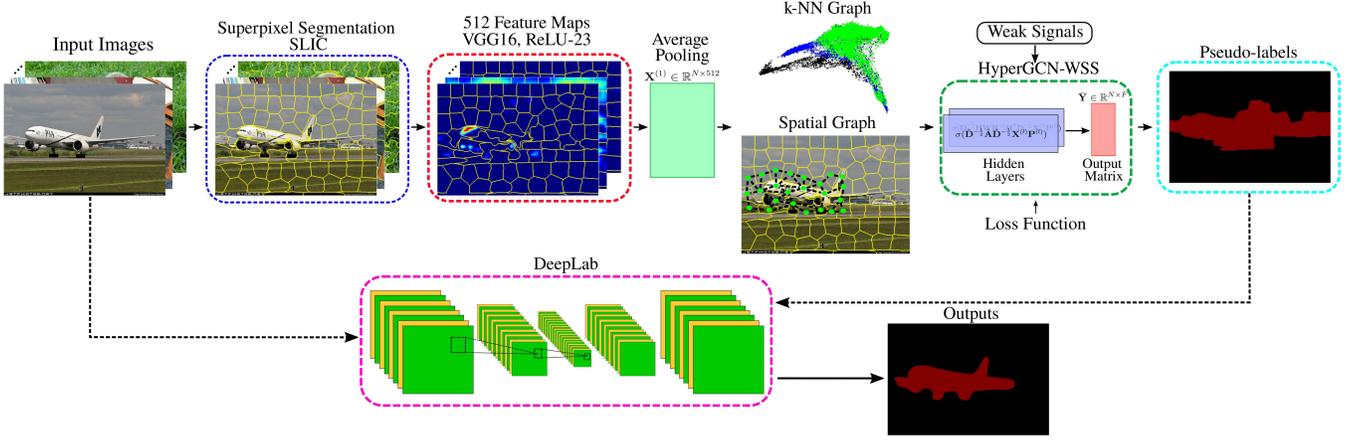}
    \caption{The pipeline of HyperGCN-WSS. Our algorithm uses SLIC superpixel segmentation, VGG16 feature extraction, average pooling, spatial and k-NN graph construction, a specialized HyperGCN architecture, and a DeepLabV3+ model.}
    \label{fig:pipeline}
\end{figure*}

The main contributions of this paper are presented as follows: 1) we propose a new algorithm for WSS, 2) we show that HyperGCNs is better than GCNs for WSS, and 3) we evaluate our algorithm with two types of weak signals, showing competitive performance against previous methods.
The rest of the paper is organized as follows.
Section \ref{sec:hypergcn_seg} explains HyperGCN-WSS.
Section \ref{sec:experiments_results} introduces the experiments and results.
Finally, Section \ref{sec:conclusions} presents the conclusions.

\section{Proposed Method}
\label{sec:hypergcn_seg}

Fig. \ref{fig:pipeline} shows the pipeline of HyperGCN-WSS, where we have superpixel segmentation, feature extraction, hypergraph construction, HyperGCN, and DeepLab for segmentation.

\subsection{Preliminaries}

A simple graph $G=(\mathcal{V,E})$ is a mathematical entity where we have a set of nodes $\mathcal{V} \in \{1,\dots,N\}$ and a set of edges $\mathcal{E}=\{(i,j)\}$.
In this paper, we consider undirected and weighted graphs.
Let $\mathbf{A} \in \mathbb{R}^{N \times N}$ be the adjacency matrix of $G$ such that $\mathbf{A}(i,j) > 0$ if $(i,j) \in \mathcal{E}$, and $0$ otherwise.
Let $\mathbf{D} \in \mathbb{R}^{N \times N}$ be the diagonal degree matrix of $G$ such that $\mathbf{D}=\diag(\mathbf{A1})$, where $\mathbf{1}$ is a vector of ones with appropriate dimension, and $\diag(\cdot)$ creates a diagonal matrix from a vector.
Notice that $\mathbf{A}$ can only represent edges that connect two nodes.
A hypergraph $G_h=(\mathcal{V,E})$ is a generalization of simple graphs $G$, where the edges can connect multiple nodes.
Let $\mathbf{W} \in \mathbb{R}^{M \times M}$ be the diagonal matrix of hyperedge weights, where $\mathbf{W}(e,e)$ is the weight of the $e$th hyperedge and $M=\vert \mathcal{E} \vert$.
Let $\mathbf{H} \in \{0,1\}^{N \times M}$ be the incidence matrix of $G_h$ such that $\mathbf{H}(i,e) = 1$ if the $i$th node is incident to the edge $e$, and $0$ otherwise, \ie $\mathbf{H}(i,e) = 1$ if the node $i$ is connected by the edge $e$.
Let $\mathbf{D}_h \in \mathbb{R}^{N \times N}$ be the diagonal matrix of node degree, where $\mathbf{D}_h(i,i)=\sum_{e=1}^M \mathbf{W}(e,e)\mathbf{H}(i,e)$.
Finally, let $\mathbf{B} \in \mathbb{R}^{M \times M}$ be the diagonal matrix of hyperedge degree, where $\mathbf{B}(e,e)=\sum_{i=1}^N \mathbf{H}(i,e)$.
In this work, we use the hypergraphs to represent two kinds of relationships: 1) the spatial relationships of the nodes on each image and 2) the relationship of nodes from different images in the dataset.

\subsection{Nodes Representation and Graph Construction}

We use the SLIC superpixel segmentation \cite{achanta2012slic} method to represent the nodes in the graph $G$ (or $G_h$).
Superpixels obtain homogeneous regions from the images to have a better context for the representation.
Furthermore, the input feature description of each node is obtained with some pre-trained Convolutional Neural Network (CNN).
In the current work, we use the outputs of the $10$th ReLU layer of the VGG16 \cite{simonyan2015very} (the $23$th layer of the network), \ie we use an intermediate layer of the CNN.
The feature representation contains $512$ features maps.
Additionally, an average pooling is performed on the superpixel regions of each feature map to obtain the feature representation, \ie each node is represented with a $512$-dimensional vector.

In the current work, we construct two types of graphs: 1) spatial graphs in the superpixels of each image, and 2) k-NN graphs with $\text{k}=10$ on some embedding space.
Let $\alpha$ be the number of images in the dataset, let $\xi$ be the number of superpixels for SLIC, and let $\mu$ be the maximum number of nodes we allow for each graph ($\mu=40000$ in the experiments).
Therefore, we construct $\tau=\lfloor \frac{\alpha \times \xi}{\mu} \rceil$ graphs, where we have $\gamma = \lfloor \frac{\alpha}{\tau} \rceil$ images per graph.
For the spatial graphs, we connect all the nodes that are in the neighborhood of each superpixel as shown in Fig. \ref{fig:pipeline}.
Therefore, we create a block diagonal matrix with the $\gamma$ adjacency matrices of each image, \ie we have $\gamma$ unconnected subgraphs for each spatial graph.
For the k-NN graph, we use an embedding representation to compute the Euclidean distances.
For example, these embeddings can be intermediate outputs of a GCN or a HyperGCN.
The weights of the edges for the spatial and k-NN graphs are given by the Gaussian function $\exp{\left({-\Vert \mathbf{x}_i - \mathbf{x}_j \Vert_2^2/\sigma^2}\right)}$, where $\mathbf{x}_i$ is the embedding (or feature representation) of the $i$th node, and $\sigma$ is the standard deviation given by $\sigma = \frac{1}{\vert \mathcal{E} \vert}\sum_{(i,j) \in \mathcal{E}} \Vert \mathbf{x}_i - \mathbf{x}_j \Vert_2$.

\begin{figure*}
    \centering
    \includegraphics[width=\textwidth]{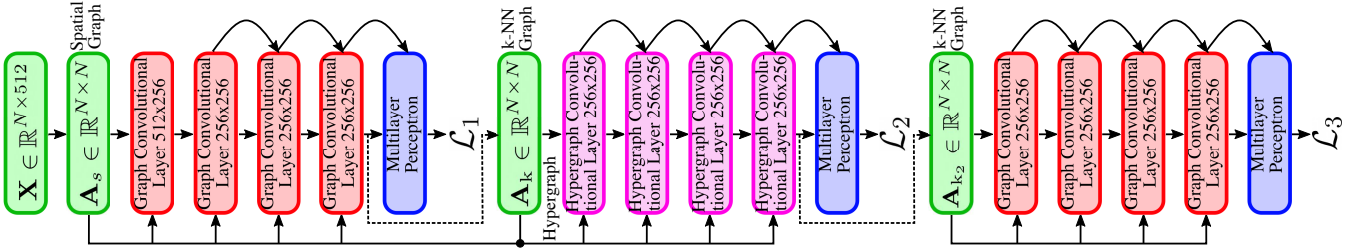}
    \caption{HyperGCN-WSS architecture with skip connections, as well as several graph and hypergraph convolutional layers. $\mathbf{X}$ is the matrix of features from VGG16. HyperGCN-WSS is trained in three stages, where we have three loss functions $\mathcal{L}_i$.}
    \label{fig:HyperGCN-WSS_architecture}
\end{figure*}

\subsection{Graph and Hypergraph Convolutional Networks}

In this paper, we use graph convolutions and hypergraph convolutions in our HyperGCN architecture.
For the graph convolutions, we use the model of Kipf and Welling \cite{kipf2017semi}.
For the hypergraph convolutions, we use the model of Bai \etal \cite{bai2021hypergraph}.

The graph convolution in \cite{kipf2017semi} is given by the following propagation rule:
\begin{equation}
    \mathbf{X}^{(l+1)}=\sigma(\tilde{\mathbf{D}}^{-\frac{1}{2}}\tilde{\mathbf{A}}\tilde{\mathbf{D}}^{-\frac{1}{2}}\mathbf{X}^{(l)}\mathbf{P}^{(l)}),
    \label{eqn:propagation_rule_GCN}
\end{equation}
where $\tilde{\mathbf{A}}=\mathbf{A}+\mathbf{I}$, $\tilde{\mathbf{D}}$ is the degree matrix of $\tilde{\mathbf{A}}$, $\mathbf{X}^{(l)}$ is the matrix of activations in layer $l$ (matrix of features or embeddings), $\mathbf{P}^{(l)}$ is the matrix of trainable weights in layer $l$, and $\sigma(\cdot)$ is an activation function.
Similarly, the hypergraph convolution in \cite{bai2021hypergraph} is given as follows:
\begin{equation}
    \mathbf{X}^{(l+1)}=\sigma(\mathbf{D}_h^{-\frac{1}{2}}\mathbf{HWB}^{-1}\mathbf{H}^{\mathsf{T}}\mathbf{D}_h^{-\frac{1}{2}}\mathbf{X}^{(l)}\mathbf{P}^{(l)}).
    \label{eqn:propagation_rule_HyperGCN}
\end{equation}

\subsection{HyperGCN Architecture}
\label{sec:HyperGCN_architecture.}

Fig. \ref{fig:HyperGCN-WSS_architecture} shows the architecture of our HyperGCN-WSS.
The input $\mathbf{X}\in \mathbb{R}^{N\times 512}$ is the matrix of features from the VGG16 network.
Each Graph Convolutional Layer (GCL) contains batch normalization \cite{ioffe2015batch}, Exponential Linear Unit (ELU) \cite{clevert2016fast} as activation function, and it could have a residual connection \cite{he2016deep} as shown in Fig. \ref{fig:HyperGCN-WSS_architecture}.
The GCLs implement the propagarion rule in \eqref{eqn:propagation_rule_GCN}.
The Hypergraph Convolutional Layers (HCLs) are similar to the GCLs, but instead of using \eqref{eqn:propagation_rule_GCN}, they implement \eqref{eqn:propagation_rule_HyperGCN}.
Our architecture also contains Multi-layer Perceptrons that classify the embedding of the GCLs or HCLs.
We use intermediate embeddings as shown by the dotted lines in Fig. \ref{fig:HyperGCN-WSS_architecture} to construct k-NN graphs.
The first k-NN graph $\mathbf{A}_{\text{k}}$ is combined with the spatial graph to create a hypergraph and the second intermediate embeddings are used to construct the k-NN graph $\mathbf{A}_{\text{k}_2}$.
We avoid the over-smoothing problem \cite{li2018deeper} by performing the training procedure in three separate steps with the loss functions $\mathcal{L}_1$, $\mathcal{L}_2$, and $\mathcal{L}_3$ \cite{chen2020iterative}.

\section{Experiments and Results}
\label{sec:experiments_results}

\subsection{Dataset and Evaluation Metrics}

HyperGCN-WSS is evaluated on PASCAL VOC 2012 \cite{everingham2015pascal} dataset for semantic segmentation.
We also use the dataset of scribbles \cite{lin2016scribblesup} and random clicks as weak signals.
PASCAL VOC 2012 has 20 semantic classes and one background category.
We use the augmented version of the PASCAL dataset provided by \cite{hariharan2011semantic}, resulting in $10582$ images in the training set, and $1449$ images in the validation set.
In this paper, we use the training set in \cite{hariharan2011semantic} for training and validation, and we leave the validation set as the test set.
We use the mean Interception over Union (mIoU) metric \cite{everingham2015pascal} for evaluation.

\subsection{Implementation Details}

HyperGCN-WSS is implemented using PyTorch with a learning rate of $0.01$ and weight decay of $5$e$-4$.
Each GCL or HCL has $256$ hidden units and a dropout rate of $50\%$.
Each stage of HyperGCN-WSS is trained for a maximum of $1000$ epochs using Adam \cite{kingma2015adam}.
For scribbles, we use $5\%$ of the scribbles for validation. 
For clicks, we use $1\%$ of the clicks for validations.
We train HyperGCN-WSS using a learning scheduler that reduces the learning rate when the loss function has stopped improving in the validation set.
Our scheduler has a reducing factor of $0.5$, patience of $25$ epochs, and a minimum learning rate of $1$e$-6$ (HyperGCN-WSS stops the learning procedure either if we reach the maximum number of epochs or if we reach the minimum learning rate).
The final activation of the Multi-layer Perceptrons are logarithmic softmax, and we use the negative log-likelihood as loss functions.

\subsection{Experiments}

In this work, we perform experiments in 1) the dataset of scribbles \cite{lin2016scribblesup} and 2) some random clicks that are given by a percentage of $N$.
The percentage of clicks is given by the set $\mathcal{M}=\{\frac{1}{32}, \frac{1}{16}, \frac{1}{8}, \frac{1}{4}\}$.
For example, for $\frac{1}{32} \in \mathcal{M}$ and $\xi=100$ number of superpixels, we have around $3.125$ random clicks per image for training.
Similarly, we analyze the impact of the number of superpixels $\xi$ in the set $\mathcal{S}=\{50,100,200,400,800\}$, for both scribbles and clicks.
We report the mIoU of the pseudo-labels in the training set for each loss function $\mathcal{L}_i~\forall~i\in \{1,2,3\}$ to assess the propagation of information of our algorithm after each stage.
We also report the mIoU after the DeepLab training in the validation set.
We do not perform an extensive search of semantic segmentation models like in \cite{pu2018graphnet,tang2018normalized} because that is not our scope.

\begin{figure*}[ht]
    \centering
    \includegraphics[width=\textwidth]{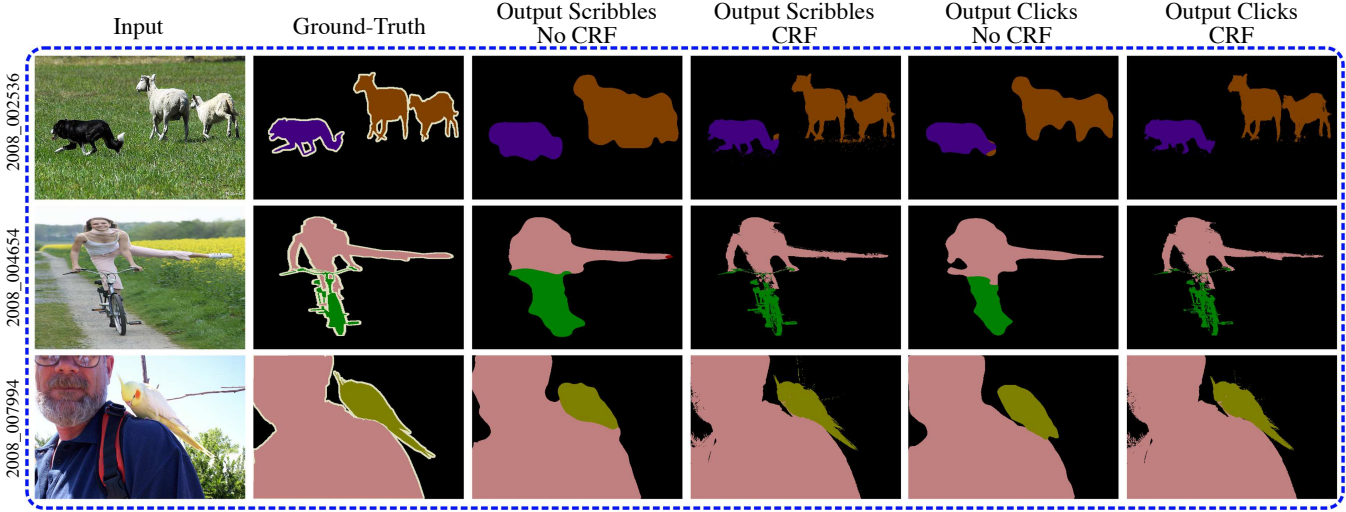}
    \caption{Some visual results on PASCAL VOC 2012 with our HyperGCN-WSS, using scribbles or clicks as weak signals.}
    \label{fig:visual_results}
\end{figure*}

\begin{table*}[]
\centering
\caption{Accuracy in mIoU (\%) in the train set of PASCAL VOC after each loss function $\mathcal{L}_i~\forall~i\in \{1,2,3\}$ in our algorithm.}
\label{tbl:mIoU_pseudo_labels}
\makebox[\linewidth]{
\scalebox{0.88}{
\begin{tabular}{r|ccc|ccc|ccc|ccc|ccc}
\hline
\multirow{2}{*}{\textbf{\begin{tabular}[c]{@{}l@{}}Weak\\ Signals\end{tabular}}} & \multicolumn{3}{c|}{$\xi=50$} & \multicolumn{3}{c|}{$\xi=100$} & \multicolumn{3}{c|}{$\xi=200$} & \multicolumn{3}{c|}{$\xi=400$} & \multicolumn{3}{c}{$\xi=800$} \\
    & $\mathcal{L}_1$ & $\mathcal{L}_2$ & $\mathcal{L}_3$ & $\mathcal{L}_1$ & $\mathcal{L}_2$ & $\mathcal{L}_3$ & $\mathcal{L}_1$ & $\mathcal{L}_2$ & $\mathcal{L}_3$ & $\mathcal{L}_1$ & $\mathcal{L}_2$ & $\mathcal{L}_3$ & $\mathcal{L}_1$ & $\mathcal{L}_2$ & $\mathcal{L}_3$ \\
\hline
Scribbles & $51.85$ & $54.49$ & $54.23$ & $50.04$ & $54.31$ & $\textbf{54.51}$ & $46.03$ & $49.72$ & $50.72$ & $39.86$ & $44.65$ & $45.51$ & $32.42$ & $37.99$ & $39.51$ \\
Clicks $\frac{1}{32}$ & $41.35$ & $42.13$ & $42.05$ & $41.43$ & $42.61$ & $42.30$ & $40.46$ & $43.06$ & $42.97$ & $39.36$ & $43.76$ & $\textbf{44.28}$ & $32.56$ & $37.91$ & $39.87$ \\
Clicks $\frac{1}{16}$ & $45.59$ & $46.78$ & $46.70$ & $46.06$ & $47.98$ & $47.72$ & $44.81$ & $49.17$ & $\textbf{49.19}$ & $39.84$ & $45.84$ & $46.98$ & $33.60$ & $38.77$ & $41.48$ \\
Clicks $\frac{1}{8}$ & $50.04$ & $51.71$ & $51.48$ & $51.08$ & $\textbf{54.21}$ & $54.15$ & $47.94$ & $52.98$ & $53.18$ & $41.33$ & $46.48$ & $47.98$ & $33.16$ & $38.76$ & $41.24$ \\
Clicks $\frac{1}{4}$ & $53.44$ & $55.76$ & $55.52$ & $53.63$ & $\textbf{57.60}$ & $57.51$ & $49.74$ & $54.51$ & $55.26$ & $41.42$ & $46.92$ & $48.82$ & $34.19$ & $37.49$ & $40.07$ \\
\hline
\end{tabular}
}
}
\end{table*}

\begin{table}[]
\centering
\caption{Accuracy of HyperGCN-WSS and other methods in the validation set of PASCAL VOC. S: Scribbles. C: Clicks.}
\label{tbl:comparisons}
\begin{tabular}{r|ccc}
\hline
Method            & \begin{tabular}[c]{@{}c@{}}Weak\\ Signal\end{tabular} & CRF & mIoU (\%) \\
\hline
ScribbleSup \cite{lin2016scribblesup} & S & \checkmark & $63.1$ \\
RAWKS \cite{vernaza2017learning} & S & \checkmark & $61.4$ \\
NormalizedCutLoss \cite{tang2018normalized} & S & - & $60.5$ \\
GraphNet \cite{pu2018graphnet} & S & - & $63.3$ \\
\hline
HyperGCN-WSS (\textcolor{red}{ours}) & S & - & $65.3$ \\
HyperGCN-WSS (\textcolor{red}{ours}) & C & - & $\textbf{65.4}$ \\
\hline
\end{tabular}
\end{table}

\subsection{Results and Discussions}

Fig. \ref{fig:visual_results} shows some visual results of HyperGCN-WSS before and after applying Conditional Random Field (CRF) \cite{krahenbuhl2011efficient} for visualization purposes.
Similarly, Table \ref{tbl:mIoU_pseudo_labels} shows the mIoU of the pseudo-labels after each loss function $\mathcal{L}_i~\forall~i\in \{1,2,3\}$ in the training set of PASCAL VOC with scribbles and clicks, \ie we have information of different parts of our architecture.
Notice that we do not use the full labeled annotation of the training set of PASCAL VOC, so Table \ref{tbl:mIoU_pseudo_labels} shows how well the information is propagated to the other nodes in the graph.
We notice that there is a gap in performance between using the GCN alone and the HyperGCN.
For example, in scribbles and clicks, there is a gap of around $4\%$ between $\mathcal{L}_1$ and $\mathcal{L}_2$.
The best results for each weak signal in Table \ref{tbl:mIoU_pseudo_labels} (in \textbf{bold}) correspond to low values of superpixels $\xi$.
The reason is that the dimensions of the features maps of the VGG16 in the $23$ layer are around $8$ times smaller than the original image. 
Therefore, having big values of $\xi$ means smaller superpixels, which are hard to adequately represent with deep layers of CNNs.
Finally, Table \ref{tbl:comparisons} shows the comparison of HyperGCN-WSS with previous methods.
We did not report results with CRF post-processing due to space constraints.
Our algorithm shows competitive performance against the other methods.

\section{Conclusions}
\label{sec:conclusions}

In this work, we introduced a new HyperGCN-WSS algorithm.
Our algorithm is composed of SLIC superpixel segmentation, CNN feature extraction, hypergraph construction, a specialized HyperGCN architecture, and the DeepLabV3+ model.
This new HyperGCN architecture combines graph and hypergraph convolutions.
HyperGCN-WSS used spatial graphs constructed in the neighborhood of the superpixels, and k-NN graphs constructed in some embedding representation.
We showed that using hypergraph convolutions is better than using graph convolutions alone.
Similarly, our algorithm showed competitive performance against previous methods.

\bibliographystyle{IEEEbib}
\bibliography{bibfile}

\end{document}